\documentclass{article}

\usepackage{PRIMEarxiv}

\usepackage[utf8]{inputenc} % allow utf-8 input
\usepackage[T1]{fontenc}    % use 8-bit T1 fonts
\usepackage{hyperref}       % hyperlinks
\usepackage{url}            % simple URL typesetting
\usepackage{booktabs}       % professional-quality tables
\usepackage{amsfonts}       % blackboard math symbols
\usepackage{nicefrac}       % compact symbols for 1/2, etc.
\usepackage{microtype}      % microtypography
\usepackage{lipsum}
\usepackage{fancyhdr}       % header
\usepackage{graphicx}       % graphics
\usepackage{xcolor,colortbl}
\usepackage{multirow}
\usepackage{makecell}
\usepackage{xspace}
\usepackage{authblk}
\usepackage{ifthen}  % Provides \ifthenelse command
\usepackage{pgffor} % Required for the \foreach command
\usepackage[symbol]{footmisc}
\usepackage{amsmath,epsfig}
\usepackage{color}
\graphicspath{{media/}}     % organize your images and other figures under media/ folder

\makeatletter
\DeclareRobustCommand\onedot{\futurelet\@let@token\@onedot}
\def\@onedot{\ifx\@let@token.\else.\null\fi\xspace}

\makeatother

\title{COCO is ``ALL'' You Need for Visual Instruction Fine-tuning}

\author[1]{Xiaotian Han}
\author[1, $\ddagger$]{Yiqi Wang}
\author[1]{Bohan Zhai}
\author[1]{Quanzeng You}
\author[1]{Hongxia Yang}

\affil[1]{ByteDance Inc., \{xiaotian.han, quanzeng.you, hx.yang\}@bytedance.com}

\begin{document}
\maketitle
\begin{abstract}
Multi-modal Large Language Models (MLLMs) are increasingly prominent in the field of artificial intelligence. 
Visual instruction fine-tuning (IFT) is a vital process for aligning MLLMs' output with user's intentions. 
High-quality and diversified instruction following data is the key to this fine-tuning process. 
Recent studies propose to construct visual IFT datasets through a multifaceted approach: transforming existing datasets with rule-based templates, employing GPT-4 for rewriting annotations, and utilizing GPT-4V for visual dataset pseudo-labeling. 
LLaVA-1.5 adopted similar approach and construct LLaVA-mix-665k, which is one of the simplest, most widely used, yet most effective IFT datasets today.
Notably, when properly fine-tuned with this dataset, MLLMs can achieve state-of-the-art performance on several benchmarks. 
However, we noticed that models trained with this dataset often struggle to follow user instructions properly in multi-round dialog. 
In addition, tradition caption and VQA evaluation benchmarks, with their closed-form evaluation structure, are not fully equipped to assess the capabilities of modern open-ended generative MLLMs.
This problem is not unique to the LLaVA-mix-665k dataset, but may be a potential issue in all IFT datasets constructed from image captioning or VQA sources, though the extent of this issue may vary. 
We argue that datasets with diverse and high-quality detailed instruction following annotations are essential and adequate for MLLMs IFT. 
In this work, we establish a new IFT dataset, with images sourced from the COCO dataset along with more diverse instructions. 
Our experiments show that when fine-tuned with out proposed dataset, MLLMs achieve better performance on open-ended evaluation benchmarks in both single-round and multi-round dialog setting. 
\end{abstract}

% keywords can be removed

%\footnotetext[1]{Project lead}
%\footnotetext[2]{Directional lead}
\footnotetext[3]{Work done during internship at ByteDance.}

\section{Introduction}
\label{sec:intro}

The remarkable progress in Large Language Models (LLMs), \textit{e.g.} GPT-4~\cite{openai2023gpt4}, LLaMA~\cite{touvron2023llama}, Mistral~\cite{jiang2023mistral}, and others, has paved the way for the emergence of Multi-modal Large Language Models (MLLMs). 
Pre-trained MLLMs often struggle with poor generalization to new tasks and limited capability in following user instructions.
The process of Visual Instruction Fine-tuning (IFT)~\cite{liu2023visual} plays a crucial role in adapting the capabilities of advanced pre-trained models. 
This fine-tuning ensures these models align more accurately with the specific intentions and needs of their users.
\begin{figure}[!htbp]
\centering
\includegraphics[width=0.6\textwidth]{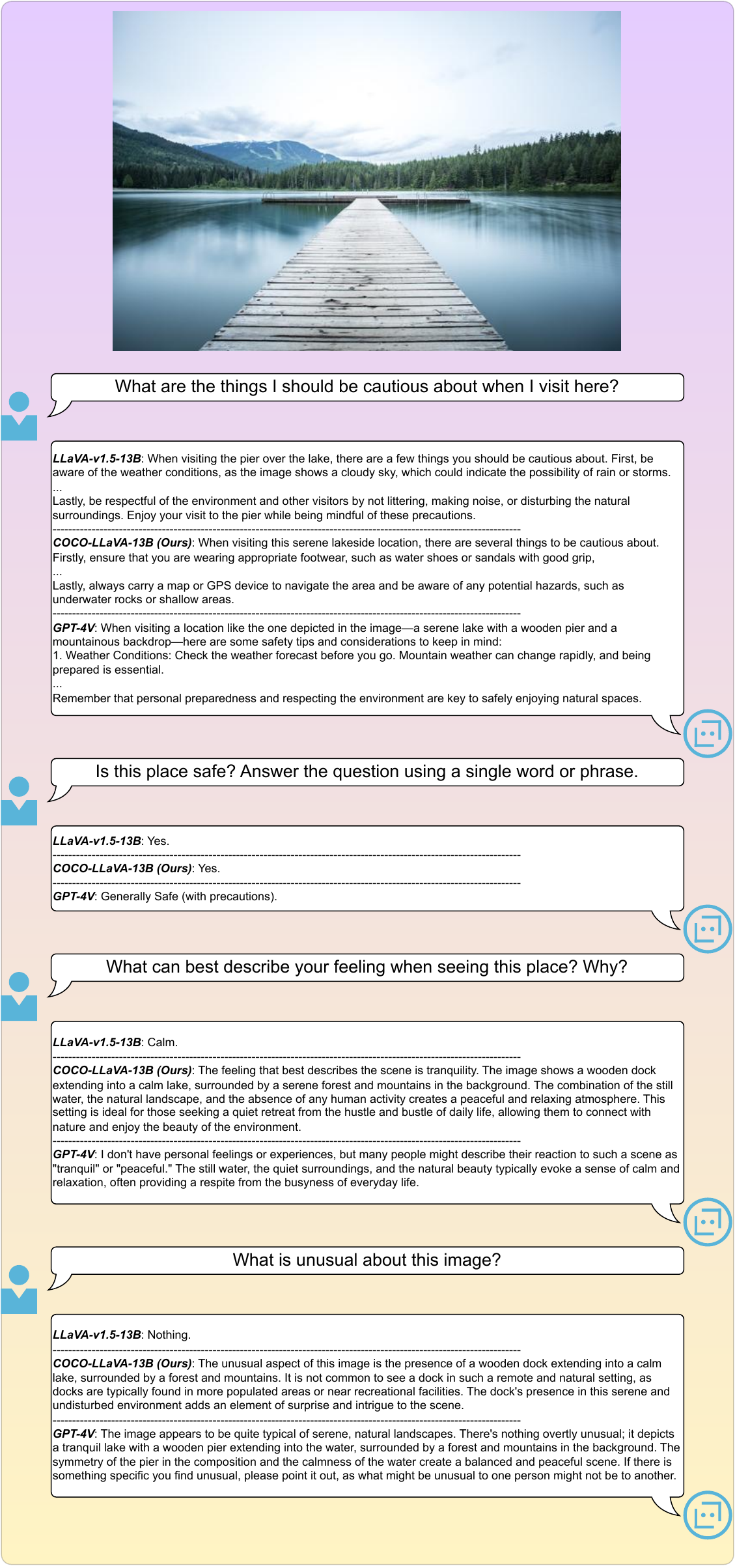}
\caption{Demonstration of different models' responses under multi-round dialog setting.}
\label{fig:llava_multi_round}
\end{figure}

IFT requires high-quality datasets, which should include multiple rounds of  \textsc{⟨instruction, response⟩} pairs within each training sample. 
A visual instruction-response pair usually includes a clearly described question or instruction interleaved with images and a detailed and accurate response that answers the question following user's instruction. 
Several Visual IFT datasets have been proposed in recent studies, \textit{e.g.} InstructBlip~\cite{dai2023instructblip}, LLaVA~\cite{liu2023visual}, MIMIC-IT~\cite{li2023mimicit}, LLaVAR~\cite{zhang2023llavar}, SVIT~\cite{zhao2023svit}, LAMM~\cite{yin2023lamm}, LVIS-INSTRUCT4V~\cite{wang2023believe}, \textit{etc}.

Traditional vision-language models evaluation benchmarks include captions and VQAs, \textit{e.g.} NoCaps~\cite{agrawal2019nocaps}, COCO~\cite{chen2015microsoft}, GQA~\cite{hudson2019gqa}, VQAv2~\cite{balanced_vqa_v2}, VizWiz~\cite{gurari2018vizwiz}, \textit{etc.} 
These benchmarks usually calculate closed-form metrics, such as CIDEr~\cite{vedantam2015cider} and accuracy by exactly answer matching. 
However, these benchmarks usually require short descriptions of image or short answers with one or two words to exactly match with ground truth, which is not suitable for generative MLLMs. 
Recently, more benchmarks designed specifically for MLLMs have been constructed, like MME~\cite{fu2023mme}, SeedBench~\cite{li2023seedbench2}, MMMU~\cite{yue2023mmmu} for multiple choice question evaluation and MM-Vet~\cite{yu2023mmvet}, InfiMM-Eval~\cite{han2023infimmeval} for open-ended QA evaluation. 

We notice that current instruction fine-tuned models may have issues in multi-round conversations. 
Fig.~\ref{fig:llava_multi_round} shows the issues with LLaVA-1.5~\cite{liu2023improved}, which is one of the most popular open-sourced MLLMs.
Upon encountering an instruction such as ``Answer the question with a short phrase.'', the intrinsic bias to which the model overfitted during the IFT stage is triggered, leading to responses consisting of single words, even when explicitly asked to provide longer answers.
We believe that the IFT stage is more concerned with aligning with user intent rather than injecting knowledge, a task that should be addressed during the pre-training stage.
Thus, IFT datasets should contain more high-quality instructions and responses rather than more training samples.
To validate this hypothesis, we analyze various visual instruction tuning datasets, extract and merge instruction annotations centered on images from MSCOCO~\cite{lin2015microsoft} and Visual Genome~\cite{krishna2016visual}.
We then retrain LLaVA-1.5~\cite{liu2023improved} with our proposed IFT dataset, evaluate it on open-ended benchmarks, and design a protocol for evaluating under multi-round dialog setting. In summary, our main contributions are:
\begin{enumerate}
    \item Constructing a COCO-image-centric visual instruction fine-tuning dataset by analyzing and merging data from various IFT dataset sources.
    \item Retraining LLaVA-1.5 with our proposed dataset, thereby outperforming the official LLaVA-1.5-13B on open-ended evaluation benchmarks.
    \item Designing a protocol to evaluate MLLMs in a multi-round dialog setting.
    \item Proving that a few images with high-quality instruction-following annotations are sufficient for IFT, and that adding more GQA or VQA datasets leads to overfitting to in-domain evaluation benchmarks.
\end{enumerate}

\section{Background}
\label{sec:background}

\subsection{Multi-modal Large Language Models}
MLLMs can be roughly categorized into two types: (1) LLMs driven tool usage systems, \textit{e.g.} MM-REACT~\cite{yang2023mmreact}, HuggingGPT~\cite{shen2023hugginggpt}, Visual ChatGPT~\cite{wu2023visual}, etc, in which LLMs serve as controller for calling expert models from different modalities and summarize their response; (2) End-to-end MLLMs, \textit{e.g.} LLaVA-1.5~\cite{liu2023improved}, Qwen-VL~\cite{bai2023qwenvl}, Flamingo~\cite{alayrac2022flamingo}, and others, which usually consist of a pretrained visual encoder, a pretrained LLM and a vision language bridge for fusing modalities. 
We mainly focus on end-to-end MLLMs in this paper.

There are different architecture designs for end-to-end MLLMs. Flamingo~\cite{alayrac2022flamingo} uses a perceiver resampler to compress visual features and then inject into LLM via gated cross attention modules. 
BLIP-2~\cite{li2023blip2} incorporates the Q-Former, adding an alignment stage to connect the frozen LLM with the visual modality. 
Mini-GPT4~\cite{zhu2023minigpt4} fine-tunes a linear projection layer to align vision and language modalities. 
LLaVA1.5~\cite{liu2023improved} integrates large-scale instruction tuning and high-resolution ViT, achieving superior results across various benchmarks.

\subsection{Visual Instruction Fine-Tuning}
Training of MLLMs usually consists of two major stages: the pre-training stage and the visual instruction fine-tuning stage.

Pretraining stage in MLLMs is usually used for aligning visual features with language features and injecting visual knowledges, such as image content understanding, OCR capability and image regional information awareness. 
MLLMs pretraining datasets can be divided into 2 types: (1) image-text pairs, like Laion~\cite{schuhmann2022laion5b}, COYO~\cite{kakaobrain2022coyo-700m}, YFCC100M~\cite{Thomee_2016}, etc; (2) images interleaved with text, \textit{e.g.} MMC4~\cite{zhu2023multimodal}, OBELICS~\cite{laurençon2023obelics}, etc. 

The visual instruction fine-tuning stage further fine-tunes the model on visual data with text or visual instructions.
In this way, MLLMs can better understand user intents and follow instructions. 
Recent works~\cite{bai2023qwenvl, wang2023cogvlm} also add an additional stage between pretraining stage and IFT stage to further inject visual knowledge, which is called continual pretraining stage. 
Currently, there are roughly three approaches being used to construct IFT datasets: (1) using rule-based templates to convert existing VQA and caption datasets into conversational format, like MIC~\cite{zhao2023mmicl}; (2) using GPT-4 to generate user-assistant conversation based on image captions, QAs or instance level information, \textit{e.g.} LLaVA-Instruct~\cite{liu2023improved}, SVIT~\cite{zhao2023svit}; 
(3) using GPT-4V to generate new captions and QAs, like ShareGPT4V~\cite{chen2023sharegpt4v}, LVIS-instruct4v~\cite{wang2023believe}, etc. 

\section{Dataset}
\label{sec:dataset}
\subsection{Data Source}
\begin{figure}[ht]
\centering
\includegraphics[width=0.7\textwidth]{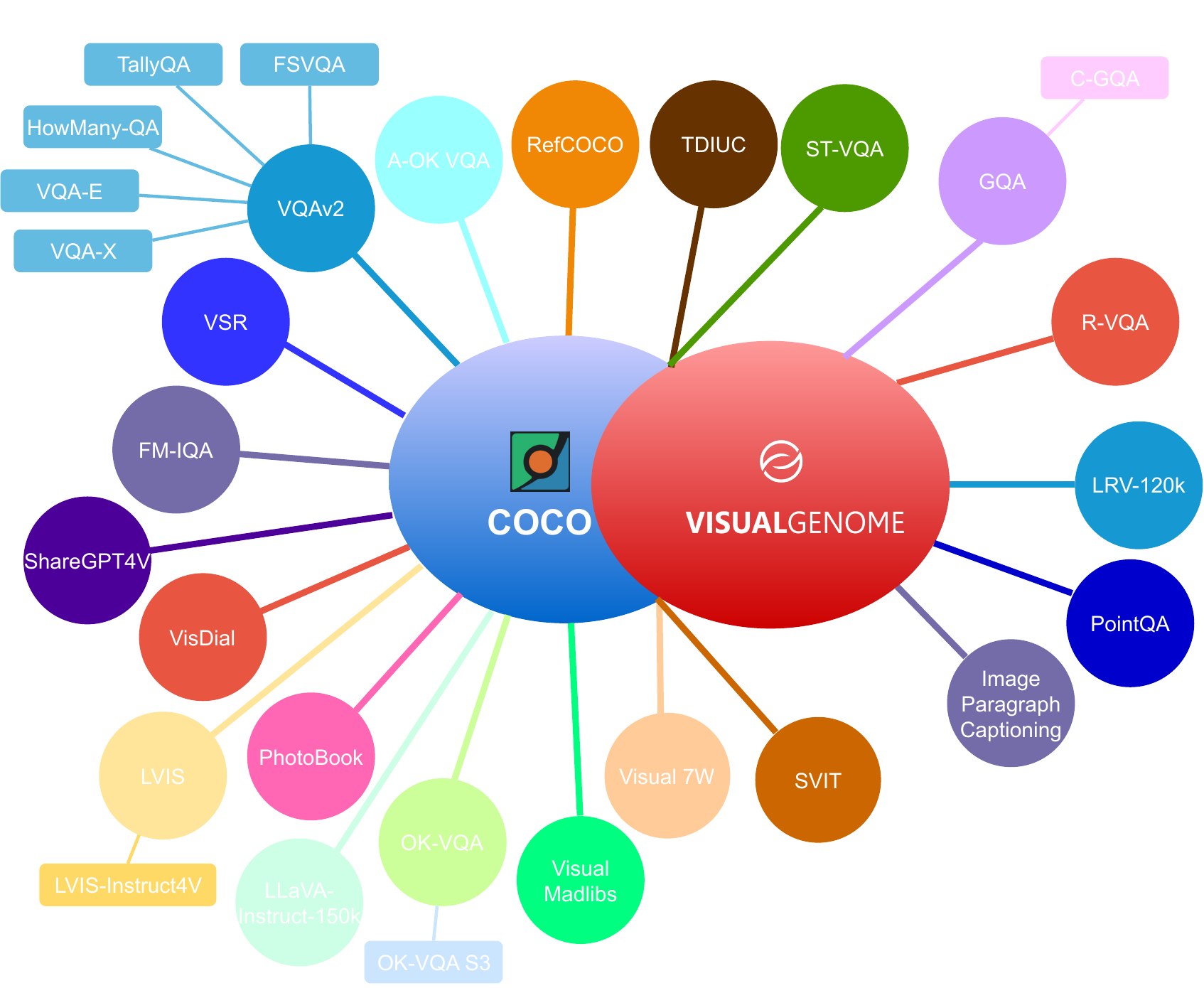}
\caption{QA and Caption datasets derived from COCO and Visual Genome.}
\label{fig:datatree}
\end{figure}

We analyze classic caption and VQA datasets and compare their dependencies in Fig.~\ref{fig:datatree}.
Our analysis suggests that COCO and Visual Genome share about half of training images. 
Both datasets contain rich manual annotations, ranging from coarse image-level captions to fine-grained object-level bounding box locations.
All other datasets shown in the figure are derived from COCO and Visual Genome.
They are either generated programmably based on original annotations or manually annotated with new QAs. 
In total, the training splits of both COCO and Visual Genome contain over 150k images, covering diverse aspects of daily usages.
The derived datasets can be categorized into 3 types: \textbf{(1) Image captions}: COCO Caption~\cite{chen2015microsoft}, Image Paragraph Captioning~\cite{krause2016paragraphs};
\textbf{(2) Visual Question Answering}: Visual Genome QA~\cite{krishna2016visual}, COCO QA~\cite{ren2015exploring}, Visual7W~\cite{zhu2016cvpr}, VQAv2~\cite{balanced_vqa_v2}, FSVQA~\cite{shin2016color}, VSR~\cite{liu2023visualspatial}, OK-VQA~\cite{okvqa}, A-OKVQA~\cite{AOKVQA}, TDIUC~\cite{kafle2017analysis}, OK-VQAS3~\cite{Jain_2021}, R-VQA~\cite{lu2018rvqa}, HowMang-QA~\cite{trott2018interpretable}, TallyQA~\cite{acharya2018tallyqa}, VQA-E~\cite{li2018vqae}, GQA~\cite{hudson2019gqa}, ST-VQA~\cite{biten2019scene}, PointQA~\cite{mani2022point}; \textbf{(3) Conversation}: VisDial~\cite{visdial_diversity}, PhotoBook~\cite{haber2019photobook}, ShareGPT4V~\cite{chen2023sharegpt4v}, LLaVA Instruct 150K~\cite{liu2023visual}, SVIT~\cite{zhao2023svit}, Sparkels~\cite{huang2023sparkles}, LAMM~\cite{yin2023lamm}, LVIS Instruct4V~\cite{wang2023believe}.
The detailed statistics of datasets can be found in Table~\ref{tab:data_statistics}.
We believe that together these datasets already cover different aspects of the images, which is sufficient to merge their annotations and build multi-round instruction fine-tuning dataset.

\subsection{Instruction Generation}
Following LLaVA~\cite{liu2023improved}, we apply a rule-based template when converting datasets into a visual instruction fine-tuning format.
The templates used are presented in Table~\ref{tab:data_templates}.

\begin{figure}[ht]
\centering
\includegraphics[width=0.8\textwidth]{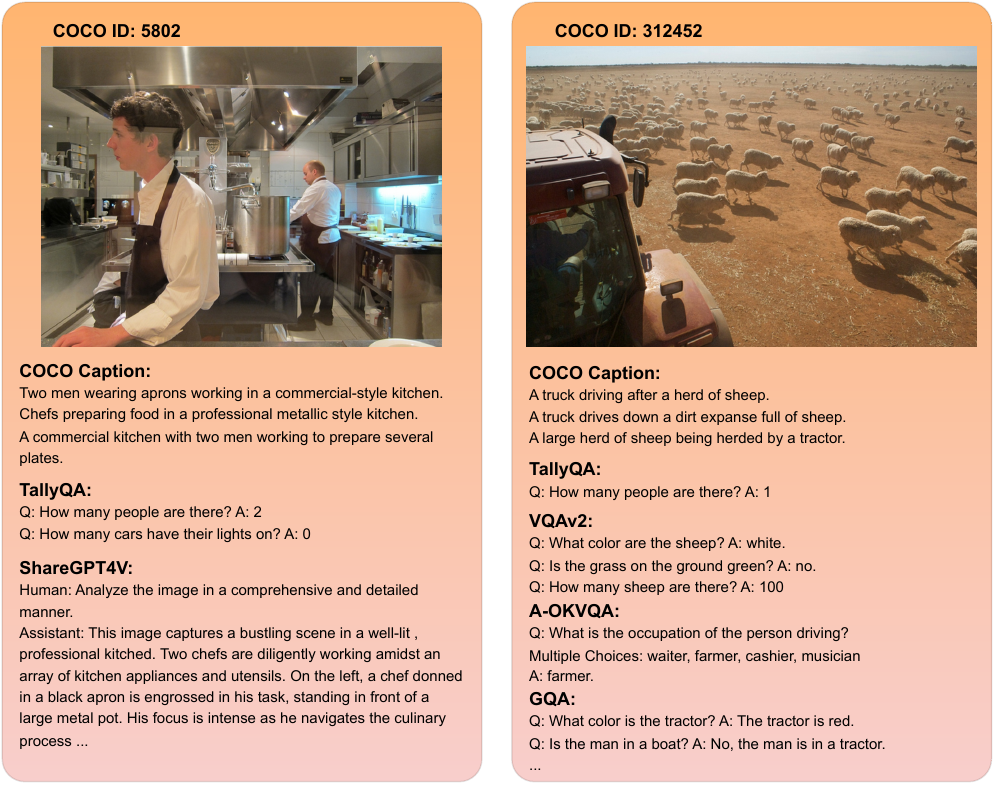}
\caption{Examples of annotations from caption and VQA datasets.}
\label{fig:data_example}
\end{figure}

Caption datasets, such as COCO Caption and Flickr30k~\cite{plummer2016flickr30k}, typically feature coarse-grained image descriptions, while VQA datasets usually have answers consisting of 1-2 words or a short phrase, as exemplified in Fig.~\ref{fig:data_example}.
Short captions and short answers are easier for evaluation using traditional metrics like CIDEr and accuracy, however, may introduce bias if used for visual instruction fine-tuning. 
Advanced prompting methods, like chain-of-thought~\cite{wei2023chainofthought} capabilities, may be undermined by these short responses.
On the other hand, short captions may also force model to ignore image details. 
Thus, we propose merging datasets based on COCO image indexes, allowing each image to feature multiple rounds of conversations related to different aspects of the image content, various levels of description, \textit{etc.}, thereby enhancing MLLMs' instruction-following capabilities.

When applying the template to QAs, we try to keep answers in full sentences, rather than as single words.
If the questions have answers with explanations, we also include these explanations as part of the model's response. 
Eventually, we obtained a training dataset comprising a total of $118,000$ samples.

\begin{table}[ht]
\begin{center}
\caption{Templates used for converting datasets into conversational IFT format} \label{tab:data_templates}
\resizebox{0.7\textwidth}{!}{
    \begin{tabular}{ll}
      \hline
      \rowcolor{gray!20} Datasets & Template \\
      \hline
      COCO Caption~\cite{chen2015microsoft} & Describe the content of this image in 20 words.  \\
      \hline
      Image Paragraph Captioning~\cite{krause2016paragraphs} & Describe the image in one paragraph. \\
      \hline
      \makecell[l]{COCO QA~\cite{ren2015exploring} \\ Visual Genome QA~\cite{krishna2016visual} \\ VQAv2~\cite{balanced_vqa_v2} \\ OK-VQA~\cite{okvqa} \\ ST-VQA~\cite{biten2019scene}} & \{question\} Answer the question with a short phrase. \\
      \hline
      TallyQA~\cite{acharya2018tallyqa} & \{question\} Answer the question with a number. \\
      \hline
      VQA-E~\cite{li2018vqae} & \makecell[l]{\{question\} Answer with a short phrase and provide \\  explanation for your answer.} \\
      \hline
      VSR~\cite{liu2023visualspatial} & \makecell[l]{\{question\} Please answer yes or no about whether the \\ statement about the image is true.} \\
      \hline
      A-OKVQA~\cite{AOKVQA} & \makecell[l]{\{question\} Answer with options\' letter from the given \\ choices and provide explanation for your choice.} \\
      % \hline
      % \makecell[l]{FSVQA~\cite{shin2016color} \\ VisDial~\cite{visdial_diversity} \\ ShareGPT4V~\cite{chen2023sharegpt4v} \\ LLaVA Instruct 150K~\cite{liu2023visual} \\ LVIS Instruct4V~\cite{wang2023believe} \\ Sparkels~\cite{huang2023sparkles} \\ LAMM~\cite{yin2023lamm} \\ GQA~\cite{hudson2019gqa} } & - \\
      \hline
    \end{tabular}
}
\end{center}
\end{table}

% \subsection{Statistics}
% In this section, we list dataset statistics in Table~\ref{tab:data_statistics}. We analyze and summarize majority caption, VQA and instruction tuning datasets that have overlap with COCO and Visual Genome. 

\begin{table}[ht]
\begin{center}
\caption{Detailed statistics of COCO derived datasets. Numbers are calculated on training set only.} 
\label{tab:data_statistics}
\renewcommand\arraystretch{1.4}
\resizebox{0.8\textwidth}{!}{
    \begin{tabular}{l|cc|cc|cc}
      \hline
      \rowcolor{gray!20}  & \multicolumn{2}{c|}{Total} & \multicolumn{2}{c|}{COCO subset} & \multicolumn{2}{c}{VG subset} \\
      \rowcolor{gray!20} \multirow{-2}{*}{Dataset} & \#samples & \#images & \#samples & \#images & \#samples & \#images \\
      \hline
      COCO Caption 2017~\cite{chen2015microsoft} & 591,753 & 118,287 & 591,753 & 118,287 & 245,330 & 49,038 \\
      \hline
      Image Paragraph Captioning~\cite{krause2016paragraphs} & 19,561 & 19,551 & 19,561 & 19,551 & 9,598 & 9,598 \\
      \hline
      COCO QA~\cite{ren2015exploring} & 78,736 & 46,293 & 78,736 & 46,293 & 32,669 & 19,098 \\ 
      \hline
      Visual Genome QA~\cite{krishna2016visual} & 1,445,322 & 108,077 & 727,063 & 49,038 & 1,445,322 & 108,077   \\ 
      \hline
      VQAv2~\cite{balanced_vqa_v2} & 443,757 & 82,783 & 443,757 & 82,783 & 179,254 & 33,848  \\ 
      \hline
      OK-VQA~\cite{okvqa} & 9,009 & 8,998 & 9,009 & 8,998 & 3,670 & 3,664 \\ 
      \hline
      ST-VQA~\cite{biten2019scene} & 26,074 & 18,897 & 21,311 & 8,909 & 9,182 & 6,982 \\
      \hline
      TallyQA~\cite{acharya2018tallyqa} & 249,318 & 132,981 & 238,056 & 99,576 & 48,429 & 33,405 \\
      \hline
      VQA-E~\cite{li2018vqae} & 181,298 & 72,680 & 181,298 & 72,680 & 73,594 & 29,719  \\
      \hline
      VSR~\cite{liu2023visualspatial} & 10,972 & 6,259 & 10,566 & 6,011 & 3,638 & 2,076 \\
      \hline
      A-OKVQA~\cite{AOKVQA} & 17,056 & 16,540 & 17,056 & 16,540 & 17,056 & 16,540 \\
      \hline
      FSVQA~\cite{shin2016color} & 662,462 & 82,783 & 662,462 & 82,783 & 270,871 & 33,848 \\ 
      \hline
      VisDial~\cite{visdial_diversity} & 123,287 & 123,287 & 118,287 & 118,287 & 49,038 & 49,038 \\ 
      \hline
      ShareGPT4V~\cite{chen2023sharegpt4v} & 102,025 & 87,300 & 50,027 & 50,012 & 20,662 & 20,654 \\ 
      \hline
      LLaVA Instruct 150K~\cite{liu2023visual} & 157,712 & 81,479 & 157,712 & 81,479 & 64,609 & 33,323 \\ 
      \hline
      LVIS Instruct4V~\cite{wang2023believe} & 222,711 & 106,944 & 222,711 & 106,944 & 88,438 & 44,259 \\ 
      \hline
      LAMM~\cite{yin2023lamm} & 185,892 & 131,405 & 107,474 & 54,100 & 101,779 & 49,034 \\ 
      \hline
      GQA~\cite{hudson2019gqa} & 943,000 & 72,140 & 454,922 & 34,411 & 943,000 & 72,140 \\
      \hline
    \end{tabular}
}
\end{center}
\end{table}

\section{Experiments}
\label{sec:experiments}

%Though LLaVA-1.5 achieved good performances across multiple benchmarks, we noticed that it performs poorly in multi-round dialog setting, due to the bias introduced when constructing IFT dataset LLaVA-mix-665k. 
Although LLaVA-1.5 achieved good performance across multiple benchmarks, we noticed it performs poorly in a multi-round dialog setting. This issue is attributed to the bias introduced during the construction of the IFT dataset LLaVA-mix-665k.
This dataset contains over 300,000 samples from GQA and Visual Genome (VG).
As discussed in Sec.~\ref{sec:dataset}, these VQA samples have short responses. 
We hypothesize that such a high number of VQA samples can only help with boosting scores for in-domain evaluation sets, such as GQA, VQAv2, TextVQA, but they cannot contribute to open-ended MLLM evaluation benchmarks, and may even hurt MLLMs' multi-round dialog capability without careful conversion.  

To validate our hypothesis, we utilized the COCO IFT dataset, as detailed in Sec~\ref{sec:dataset}, and fine-tuned LLaVA-1.5-13B from the same pretrained checkpoint.
We also design a protocol on top of MM-Vet~\cite{yu2023mmvet} and InfiMM-Eval~\cite{han2023infimmeval} to assess multi-round QA capability. 
The detailed protocol is as follows: In the first round of conversation, we feed a fixed question \textit{``What is the color of the center of this image? Answer the question using a single word or phrase."}; In the second round, we ask the model to answer the actual question of the image from the annotation. We use the same evaluation metric as original benchmarks with 2nd round answer only.

The evaluation results for MM-Vet and InfiMM-Eval, both with and without the multi-round dialog evaluation protocol, are shown in Table~\ref{tab:mmvet_res} and Table~\ref{tab:infimm_res}, respectively.

\begin{table}[h]
\begin{center}
\caption{Evaluation results on MM-Vet } \label{tab:mmvet_res}
\resizebox{0.9\textwidth}{!}{
    \begin{tabular}{lcccccccc}
      \hline
      \rowcolor{gray!20} MLLM & Multi-round & Rec & OCR & Know & Gen & Spat & Math & Total \\
      \hline
        \multirow{2}{*}{LLaVA-1.5-13B} & Yes & 31.3 & 25.7 & 7.5 & 9.3 & 33.3 & 11.2 & 29.2 \\
                                    & No &40.3 & 28.3 & 22.6 & 23.9 & 34.9 & 7.7 & 36.0 \\
        \hline
        \multirow{2}{*}{LLaVA-COCO-13B (Ours)} & Yes & 41.2 & 27.9 & 27.4 & 25.4 & 31.1 & 15.0 & 37.5 \\
                                                & No & 39.3 & 28.5 & 24.6 & 24.4 & 29.1 & 11.2 & 35.4 \\
        % \multirow{2}{*}{GPT-4V} & Yes & 58.1 & 71.7 & 47.0 & 51.0 & 70.8 & 65.4 & 63.1 \\
      \hline
    \end{tabular}
}
\end{center}
\end{table}

\begin{table}[h]
\begin{center}
\caption{Evaluation results on InfiMM-Eval } \label{tab:infimm_res}
\resizebox{0.9\textwidth}{!}{
    \begin{tabular}{lccccc}
      \hline
      \rowcolor{gray!20} Model & Multi-round & Deductive & Abductive & Analogical & Overall \\
      \hline
        \multirow{2}{*}{LLaVA-1.5-13B} & Yes & 30.64 & 27.12 & 20.83 & 28.22 \\
                                    & No & 30.94 & 47.91 & 24.31 & 32.62 \\
        \hline
        \multirow{2}{*}{LLaVA-COCO-13B (Ours)} & Yes & 36.43 & 47.12 & 28.47 & 36.77 \\
                                    & No & 36.27 & 44.70 &  24.72 & 35.56 \\
      \hline
    \end{tabular}
}
\end{center}
\end{table}

%Training with our proposed COCO-centric IFT dataset, the model shows similar performance in single round and multi-round conversation setting on both evaluation benchmarks. 
After training with our proposed COCO-centric IFT dataset, the model demonstrates similar performance in both single-round and multi-round conversation settings on both evaluation benchmarks.
However, the original LLaVA-1.5-13B exhibits a significant drop in performance after adopting the multi-round evaluation protocol.
On MM-Vet, the two subsections where the model's performance drops the most are ``Generation'' and ``Knowledge''. This aligns with our observation that the model outputs only a single word after being prompted with the first-round question.
On InfiMM-Eval, the section where the model shows the most significant drop is in ``Abductive Reasoning'' questions, which also require the model to generate reasons for the results mentioned in the question.
These experiments support our hypothesis that high-quality and diversified instructions focused solely on COCO images are sufficient for IFT.

\section{Conclusion and Limitations}
\label{sec:conclusion_and_limitation}

\textbf{Conclusion.} In this paper, we uncover an issue caused by overfitting during IFT training in current open-source MLLM models. 
This overfitting leads to a degradation in performance in multi-round dialog settings.
We construct an IFT dataset by simply merging datasets with COCO images.
Experiments show that models trained with our dataset demonstrate better instruction-following ability and achieve equal or better performance on open-ended evaluation benchmarks.
The results suggest that the COCO dataset is ``all'' you need for visual IFT.
We call for more comprehensive research to better understand IFT dataset construction, better evaluation benchmarks for modern open-ended MLLMs rather than traditional caption and VQA benchmarks.

\noindent\textbf{Limitations.} Despite the simplicity of proposed IFT dataset, a few limitations must be acknowledged. 
First, we simply merge annotations of the same image into a multi-round conversation without considering their order and logical relationship.
Second, multiple images interleaved with text samples are rare in our datset. 
Third, there is a need for specialized multi-round, open-ended benchmarks for MLLMs evaluation.

\newpage
%Bibliography
\bibliographystyle{unsrt}  
\bibliography{references}  

\end{document}